\documentclass[10pt,twocolumn,letterpaper]{article}

\usepackage{cvpr}
\usepackage{times}
\usepackage{epsfig}
\usepackage{graphicx}
\usepackage{amsmath}
\usepackage{amssymb}
\usepackage{algorithm}
\usepackage{url}
\usepackage{algorithmic}

\usepackage[breaklinks=true,bookmarks=false]{hyperref}

\cvprfinalcopy 

\def\cvprPaperID{****} 

\begin{document}

\title{TextMountain: Accurate Scene Text Detection via Instance Segmentation}

\author{Yixing Zhu, Jun Du\\
  National Engineering Laboratory for Speech and Language Information Processing\\ University of Science and Technology of China\\
  Hefei, Anhui, China\\
  {\tt\small zyxsa@mail.ustc.edu.cn, jundu@ustc.edu.cn}
}

\maketitle

\begin{abstract}
In this paper, we propose a novel scene text detection method named TextMountain. The key idea of TextMountain is making full use of border-center information. Different from previous works that treat center-border as a binary classification problem, we predict text center-border probability (TCBP) and text center-direction (TCD). The TCBP is just like a mountain whose top is text center and foot is text border. The mountaintop can separate text instances which cannot be easily achieved using semantic segmentation map and its rising direction can plan a road to top for each pixel on mountain foot at the group stage. The TCD helps TCBP learning better. Our label rules will not lead to the ambiguous problem with the transformation of angle, so the proposed method is robust to multi-oriented text and can also handle well with curved text. In inference stage, each pixel at the mountain foot needs to search the path to the mountaintop and this process can be efficiently completed in parallel, yielding the efficiency of our method compared with others. The experiments on MLT, ICDAR2015, RCTW-17 and SCUT-CTW1500 databases demonstrate that the proposed method achieves better or comparable performance in terms of both accuracy and efficiency. It is worth mentioning our method achieves an F-measure of $76.85\%$ on MLT which outperforms the previous methods by a large margin. Code will be made available.

\end{abstract}

\section{Introduction}

Recently, scene text detection has become increasingly popular, many researchers pay more attention to it due to its wide applications such as image and video retrieval, automatic driving and scene text translation. Although the academia has been studying for many years, scene text detection is still challenging because of great varieties in shape, size, angle and complex backgrounds.

Most of Traditional scene text detection methods analyze text based on character-based structural features. In recent years, benefitting from deep learning, many methods adopt deep convolutional neural network (CNN)~\cite{krizhevsky2012imagenet} to extract features and achieve considerable improvement. Deep learning based methods can be mainly divided into two categories. One is regression based method which generates text bounding boxes by regressing the coordinates of boxes. The other one is segmentation based method which predicts the segmentation map of text line and the key part is how to split adjacent text lines.  

\begin{figure}[t]
  \begin{center}
       \includegraphics[width=1.0\linewidth]{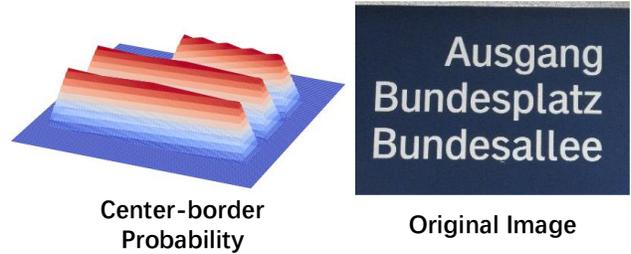}
    \end{center}
    \caption{TextMountain: The mountaintop of TCBP can separate text lines and the mountain rising direction can plan a road to mountaintop for each pixel on mountain foot at the group stage.}
  \label{overview_begin}
\end{figure}

In this study, we aim to design a concise method which can detect long text lines without the restriction of receptive field and avoid the ambiguity with the transformation of angle. Besides these we also attempt to accelerate the post-processing step to make our method more efficient. To achieve these goals, our approach follows two principles. First, we do not force the network to treat the text line as a whole object. Second, we will not define the order of vertexes and sides for a polygon which may lead to ambiguity. In particular, we utilize a fully convolutional network (FCN)~\cite{long2015fully} model to generate segmentation maps for predicting maps of text score (TS), text center-border probability (TCBP) and text center-direction (TCD). As shown in Fig.~\ref{overview_begin}, we predict a TCBP where the values for the border/center pixels is 0/1 and the values of other pixels gradually decay from center to border, the rising direction of TCBP can be used to group pixels. This is different from previous methods to treat center-border as a binary classification problem. Besides these, we also predict TCD which is not necessary in inference but can make probability map learn better. Finally, benefitting from the parallel computation of our GPU implementation, the group process consumes very little time.

The contribution of this paper are four-fold:

\begin{itemize}

\item We propose a novel method named TextMountain which is composed of TS, TCBP and TCD. Our experiments prove that TCBP can well separate text lines and group them while TCD can help TCBP learning better.

\item The proposed method can well and efficiently handle long, multi-oriented and curved scene text.

\item We propose a group algorithm using parallel computing which significantly accelerates the post-processing in inference stage.

\item The proposed method achieves state-of-the-art or comparable results in both quadrangle and curved polygon labeled datasets.

\end{itemize}

\section{Related Work}

Traditional text detection methods mainly use extremal region, border information or character's morphological information to locate text such as Stroke Width Transform (SWT)~\cite{epshtein2010detecting} and Maximally Stable Extremal Regions (MSER)~\cite{neumann2012real,zamberletti2014text}. With the emergence of deep learning, many methods try using deep neural nets to solve this problem and greatly exceed traditional methods on both performance and robustness. The deep learning based methods can be roughly divided into two categories: regression based method and segmentation based method.

Regression based method: Many text detection methods design their systems based on object detection methods and make corresponding improvements to the particularity of text.~\cite{liao2017textboxes} improves~\cite{liu2016ssd} by adjusting the aspect ratios of anchor and then setting them with vertical offsets, and adopts irregular $1 \times 5$ convolutional filters for long text lines. Different from traditional horizontal rectangle labeled object detection method,~\cite{ma2018arbitrary} proposes rotation region proposal networks which generate rotated rectangles in region proposal step,~\cite{jiang2017r2cnn} still proposes horizontal rectangles in region proposal step but regresses a rotated rectangle in R-CNN step.~\cite{liao2018textboxes++} adapts~\cite{liao2017textboxes} for multi-oriented scene text which regresses four vertexes of target box and also uses recognition information to refine detection results.~\cite{liao2018rotation} classifies text line and regresses its location with different feature which achieves significant improvement on oriented text line.~\cite{he2017deep} and~\cite{zhou2017east} investigate to generate shrinked text line segmentation map then regress text sides or vertexes on text center. Although most of scene text lines are quadrangle, there are also curved texts in a natural scene and quadrangle label may lead to background noises for these texts. Considering this,~\cite{yuliang2017detecting} proposes a dataset named SCUT-CTW1500 whose text line is labeled with polygon and a novel method called CTD which regresses multiple points on text lines, a TLOC component is also proposed to learn the correlation between points. But regressing both x and y coordinates of multiple points will increase redundancy,~\cite{zhu2018sliding} slides a line along horizontal box, then only regresses x or y coordinates of intersection between sliding lines and text polygon. In order to make recognition easier,~\cite{long2018textsnake} proposes TextSnake which can effectively represent text lines in curved form and straighten curved text lines.

\begin{figure*}
  \begin{center}
       \includegraphics[width=1.0\linewidth]{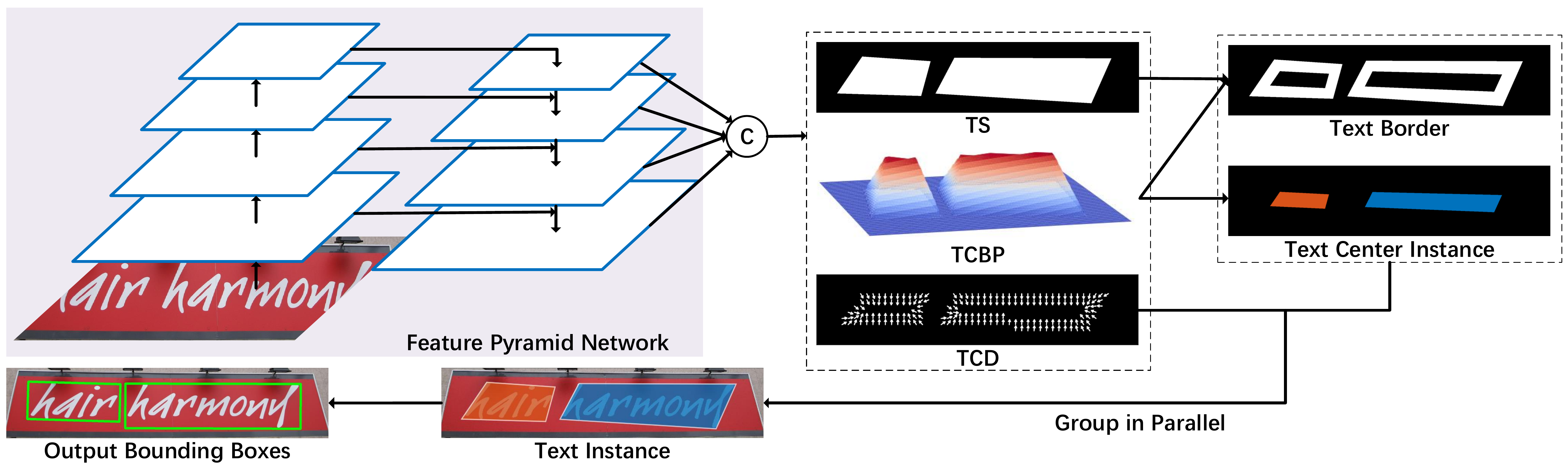}
    \end{center}
    \caption{The pipeline of our system: the Feature Pyramid Network (FPN) with feature fusion network, output (TS, TCBP and TCD) and post-processing. 
    }
  \label{pipeline}
\end{figure*}
Segmentation based method: FCN~\cite{long2015fully} is widely used to generate text segmentation map. The key point of these methods is how to split adjacent text lines.~\cite{zhang2016multi} predicts segmentation map of text regions and centroid of each character, then combines these information to group text lines.~\cite{he2016accurate,he2017multi,wu2017self} use text center or text border map to separate text lines and accelerate post-processing.~\cite{deng2018pixellink} defines border on pixel level with 8-direction and uses their connected-direction to separate and group text lines.~\cite{lyu2018multi} adopts corner detection to propose boxes and calculates each box's score on position-sensitive segmentation map.~\cite{li2018shape} proposes PSENet which generates various scales of shrinked text segmentation maps, then gradually expands kernels to generate final instance segmentation map which can handle well with curved text lines. There are also many methods adopt box proposal based instance segmentation network such as Mask R-CNN~\cite{he2017mask} for text detection.~\cite{dai2017fused} fuses multiple-layer feature map for RPN and RoI, finally predicts a segmentation map of text.~\cite{yao2018mask} completes both of detection and recognition tasks on mask-branch by predicting word segmentation map and character instance segmentation map, a weighted edit distance measure is proposed to find the best-matched recognition result. Besides instance segmentation methods designed for text, there are many general object instance segmentation methods.~\cite{de2017semantic} proposes a discriminative loss function that can push close pixels which belong to the same objects and distant pixels which belong to different objects.~\cite{novotny2018semi} proposes semi-convolutional embedding which can separate object instances cannot be easily separated with traditional convolutional operator and this operator can also improve performance of Mask R-CNN.~\cite{zhou2018weakly} exploits class peak responses for instance mask extraction. Different from general object instance segmentation methods, we design our system with rules of text shape which can make our system more efficient and robust because the shape of text line is more regular than general objects. It is primarily inspired by~\cite{de2017semantic}, pixels on the adjacent side of text lines have totally different features, so we use center-direction map to push close pixels which belong to the same text line and distant pixels which belong to different text lines. We also share a similar idea in~\cite{zhou2018weakly}, in our design, each text is a mountain on center-border probability map, we make a full use of mountain information in post-processing. Although our definition is based on regular shape of text line, we don't define the order of vertices to avoid the ambiguity with the transformation of angles.

\section{Method}
\subsection{Overview}
This section introduces the detail of TextMountain. The key idea of TextMountain is training a FCN network~\cite{long2015fully} adapted for text detection that outputs maps of text score, text center-border probability and text center-direction. As shown in Fig.~\ref{pipeline}, on the text border probability map, each text line is like a mountain whose peak is text center and foot is text border. Firstly, we set a threshold to generate text center instance map and text border map from TCBP and TS, then we calculate the score of each text center instance with its mean score on TS. The text center can separate text lines that cannot be easily separated in TS. Next, each pixel on text border map searches for its peak by TCBP using the rising direction or directly using the direction of TCD. Each pixel walks toward text center until it arrives at one text center then the pixel belongs to this text center. This step is very simple without complex rules which can be easily completed using parallel computation on GPU and consumes a little time.
\subsection{Network Design}

Inspired by recent works related to semantic segmentation~\cite{xiao2018unified,zhou2017scene,zhou2016semantic,li2018shape}, we design our model based on feature pyramid network (FPN)~\cite{lin2017feature}. FPN is a widely-used module in object detection and it is also used in semantic segmentaion in UPerNet~\cite{xiao2018unified}. The skeleton of our network is shown in Fig.~\ref{pipeline}. we concatenate low-level feature maps and high-level semantic feature maps then fuse these feature maps with extra $3 \times 3$, $1 \times 1$ convolution layers. The output feature map has $5$ channels for TS, TCBP and TCD. Finally, we upsample the predicting map to input image size. 

\subsection{Loss Functions}
There are three tasks in this method, they are TS, TCBP and TCD. And the overall loss function can be formulated as:
\begin{equation}
L=L_{\text{TS}}+\lambda_{1} L_{\text{TCBP}}+\lambda_{2}L_{\text{TCD}}
\end{equation}
where $L_{\text{TS}}$, $L_{\text{TCBP}}$ and $L_{\text{TCD}}$ are the loss functions of TS, TCBP and TCD, respectively. $\lambda_{1}$ and $\lambda_{2}$ are the balancing factors of the three tasks. By default, we set $\lambda_{1}$ to $5$ and $\lambda_{2}$ to $2.5$. Next, we will introduce the details of each task.

\subsection{TS}
TS classifies each pixel as text or non-text, we label pixels inside text polygon as positive samples whereas negative samples. Class-imbalance is a serious problem because most of pixels in one image are negative samples, there are many methods to solve class-imbalance problem such as class-balanced cross-entropy~\cite{xie2015holistically,zhou2017east}, proportional selection~\cite{ren2015faster} and hard negative mining~\cite{deng2018pixellink,felzenszwalb2010object}. In this paper, we use the hard negative mining which selects the most hard negative samples, the positive and negative sample ratio is set to $1/3$. The loss function in this task can be formulated as:
\begin{equation}
L_{\text{TS}}=\frac{1}{|\Omega|}\sum\limits_{\mathbf{x} \in \Omega}L_{cls}(TS_{\mathbf{x}},TS_{\mathbf{x}}^{*})
\end{equation}
where $\mathbf{x} \in \mathbb{R}^{H \times W}$ represents the two-dimensional coordinates of pixels, $\Omega$ is the set of negative pixels selected by hard negative mining adding with all positive samples, $|\Omega|$ represents the number of these pixels, $L_{cls}$ is a cross-entropy loss function, $TS_{\mathbf{x}}^{*}$ is the ground truth and $TS_{\mathbf{x}}$ is the prediction score.

\subsection{TCBP}

In recent years, most of text detection methods~\cite{he2016accurate,zhou2017east,wu2017self,deng2018pixellink,li2018shape} define center-border problem as a binary classification problem. In this paper, we treat center-border as a probability map because we think the hard decision of border or center is not always accurate. And there is more information in TCBP, the rising direction of probability growth points to text center which is useful for grouping pixels. In our design, the definition of curved text is the same with quadrangle, both of them is based on four sides of text line (a curved text line also has four sides but two of them may be curved). For simplicity, we illustrate our method on quadrangle text line and it can easily be extended to curved text line. In fact, there are many reasonable methods to compute center-border map and their results are the same when text line is a rectangle. To facilitate a simpler label rule which is easier to learn for network, we only use the vertical lines of four sides to compute labels, illustrated in Fig.~\ref{dis_cal}. Firstly, we define the height of text line as:
\begin{equation}
h_{\mathbf{x}}= \min(\|\mathbf{a}_{1}\|+\|\mathbf{a}_{3}\|,\|\mathbf{a}_{2}\|+\|\mathbf{a}_{4}\|)
\label{height_cal}
\end{equation}
where  $\mathbf{a}_{i}$ is a vector which is the vertical line of $i$-th side passing point $\mathbf{x}$ and the direction is from the intersection of side and vertical line to point $\mathbf{x}$, $\|\mathbf{a}_{i}\|$ is the length of $\mathbf{a}_{i}$. $\min(\|\mathbf{a}_{1}\|+\|\mathbf{a}_{3}\|,\|\mathbf{a}_{2}\|+\|\mathbf{a}_{4}\|)$ represents the minimum distance from point to two opposite sides which approximates the height text line, then the TCBP of $\mathbf{x}$ can be calculated as:
\begin{equation}
TCBP_{\mathbf{x}}=\frac{2 \times \min (\|\mathbf{a}_{1}\|,\|\mathbf{a}_{2}\|,\|\mathbf{a}_{3}\|,\|\mathbf{a}_{4}\|)}{h_{\mathbf{x}}}
\label{probability_cal}
\end{equation}
$TCBP_{\mathbf{x}}$ is a continuous function in the range of $[0, 1]$. The loss of TCBP can be formulated as:



%



\begin{figure}[t]
  \begin{center}
       \includegraphics[width=0.8\linewidth]{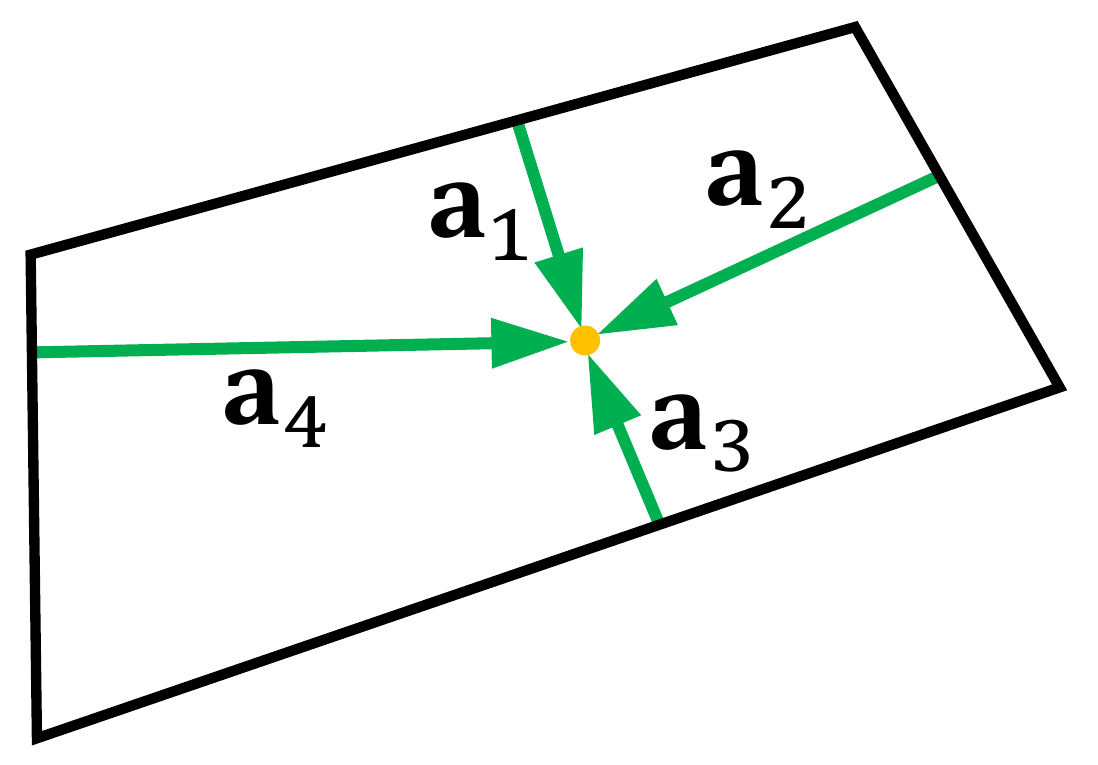}
    \end{center}
    \caption{The vertical lines of four sides. In fact, the order of $\{\mathbf{a}_{1}, \mathbf{a}_{2}, \mathbf{a}_{3}, \mathbf{a}_{4}\}$ doesn't make difference. 
    }
  \label{dis_cal}
\end{figure}

\begin{equation}
L_{\text{TCBP}}=\frac{\sum_{\mathbf{x}} TS_{\mathbf{x}}^{*}|TCBP_{\mathbf{x}}-TCBP_{\mathbf{x}}^{*}|} {\sum_{\mathbf{x}} TS_{\mathbf{x}}^{*}}
\label{mae}
\end{equation}

where $TCBP_{\mathbf{x}}$ and $TCBP_{\mathbf{x}}^{*}$ represent the prediction and ground truth of the TCBP, the activation layer of $TCBP_{\mathbf{x}}$ is sigmoid because $TCBP_{\mathbf{x}}$ is among $[0, 1]$. $TS_{\mathbf{x}}^{*}$ is the ground truth of the TS, we only calculate $L_{\text{TCBP}}$ on text region. $|TCBP_{\mathbf{x}}-TCBP_{\mathbf{x}}^{*}|$ represents the distance between prediction and ground truth. In this paper, we use L1 Loss.

\subsection{TCD}

Although there are enough information in TCBP to group pixels, we find that network can achieve better performance by predicting TCD. Each pixel on TCD will point to the center that it belongs to, it is like the fastest mountain route. The text center-direction is different from text angle defined in ~\cite{zhou2017east,long2018textsnake}. When height and width of text are equal, it is hard to define text's angle in ~\cite{long2018textsnake} and when text's angle is about $45^{\circ}$, it is hard to define text's angle in ~\cite{zhou2017east}. There is no such problem in our method. In particular, the direction vector can be formulated as:
\begin{equation}
\mathbf{v_{\mathbf{x}}}=\sum\limits_{i=1}^{4} [\frac{h_{\mathbf{x}}}{2}-\|\mathbf{a}_{i}\|]_{+} \times \frac{\mathbf{a}_{i}}{\|\mathbf{a}_{i}\|}
\label{cal_direction}
\end{equation}
where $h_{\mathbf{x}}$ is height of text line which is defined in Eq.~(\ref{height_cal}), $[z]_{+}$ represents $\max(z,0)$. We think each side has a thrust to push the point to the center and the pushing direction is the vertical line from side to point. The closer point to side, the greater the thrust is. And the thrust will be zero if the distance is longer than half of the height. Pixels on the intersection edges of two adjacent text lines have similar TCBP values  but totally different TCD values. So it can help with separating adjacent text lines. Because we only need the direction of vector, the vector is normalized as:
\begin{equation}
\mathbf{u_{\mathbf{x}}}=\frac{\mathbf{v_{\mathbf{x}}}}{\parallel \mathbf{v_{\mathbf{x}}} \parallel}
\end{equation}
L1 Loss is also used in this task and the loss function can be formulated as:
\begin{equation}
L_{\text{TCD}}=\frac{\sum_{\mathbf{x}} TS_{\mathbf{x}}^{*} (TCBP_{\mathbf{x}} < \gamma) |\mathbf{u_{\mathbf{x}}}-\mathbf{u_{\mathbf{x}}^{*}}|}{\sum_{\mathbf{x}} TS_{\mathbf{x}}^{*} (TCBP_{\mathbf{x}} < \gamma)}
\label{mae}
\end{equation}
where $TS_{\mathbf{x}}^{*}$ is the ground truth of score map, $TCBP_{\mathbf{x}}$ is the prediction of TCBP and $\gamma$ is a center threshold. $L_{\text{TCD}}$ is only valid on border region, because the TCD may increase ambiguity in center region and the TCD is only used by border pixels in inference. $\mathbf{u_{\mathbf{x}}}$ and $\mathbf{u_{\mathbf{x}}}^{*}$ are prediction and ground-truth of TCD respectively. $\mathbf{u_{\mathbf{x}}}$ is a two-dimensional vector with each dimension among $[-1,1]$, so we activate them with sigmoid then multiply them by 2 and subtract 1.

\subsection{Group in Parallel}
After we obtain TS, TCBP and TCD, a threshold $\gamma$ is set to generate mountain peak map by $TCBP_{\mathbf{x}} > \gamma$ and $\gamma$ is set to $0.6$ by default. As shown in Fig.~\ref{pipeline}, this is an instance segmentation map of text mountain peak calculated by connected domain and each peak is painted in different colors. Now we need to predict which peak each pixel on mountain foot belongs to. The calculation process is as follows: Firstly, we can generate an oriented graph by TCBP or TCD. For TCBP, we select the largest point in 8-neighbor of each pixel as the next point. For TCD, the next point is calculated as follow:
\begin{equation}
(u_{\mathbf{x}}^{1},u_{\mathbf{x}}^{2})=\mathbf{u_{\mathbf{x}}}
\end{equation}
\begin{equation}
next_{\mathbf{x}}^{(1,2)}=
\begin{cases}
1& u_{\mathbf{x}}^{(1,2)} > \cos( \frac{3}{8} \pi)\\
-1& u_{\mathbf{x}}^{(1,2)} < -\cos( \frac{3}{8} \pi)\\
0& \text{otherwise}
\end{cases}
\end{equation}
where $\mathbf{u_{\mathbf{x}}}$ is the predicted vector from TCD. We quantify the direction because there are only 8-neighbor directions, $next_{\mathbf{x}}^{(1,2)}$ is the quantified result. When the oriented graph has been generated, every pixel on mountain foot will climb to its mountain peak step by step and paint itself in peak color. Each point's direction is deterministic, so this task can be efficiently solved in parallel, all pixels can climb to the mountain at the same time. It is not necessary for every pixel to walk through all paths, when some pixels on mountain foot climb to the middle of the mountain, the pixels on the middle of mountain have climbed to the peak and they have been colored, we can directly color pixels on mountain foot with the color of pixels on middle of mountain. Actually we increase a number of computation threads, but just additionally yielding a little computation complexity. To accelerate algorithm and avoid loop, we add a block state map which indicates whether this route is block. The procedure is summarized in Algorithm~\ref{group}. This process is very fast which only needs $0.0006s$ for one image at $1280 \times 768$ resolution of ICDAR2015 dataset.

\begin{algorithm}[h]
\caption{Group in Parallel}

 {\bf Input: $C$, $S$, $D$, $B$, $ps$} \\
$C$ - text instance color map\\
$S$ - TS after setting threshold\\
$D$ - center direction map\\
$B$ - block state map, $1$ indicate this route is block.\\
$ps$ - positive pixels on border\\
$N$ - number of positive pixels\\
 {\bf Output: $C$}\\
 $C$ - text instance color map after grouping\\
\begin{algorithmic}[1]

\STATE $B[...] \leftarrow 0$
\FOR{$p \in ps$}
\STATE $p_{next} \leftarrow D[p]$
\STATE $i \leftarrow 0$
\WHILE{1}
\STATE $i \leftarrow i+1$
\IF{$B[p_{next}]==1$}
\STATE $B[p]\leftarrow 1; break$
\ENDIF

\IF{$(S[p_{next}]==0)$ or $(p_{next}==p)$ or $(i>N)$}
\STATE $B[p_{next}] \leftarrow 1; B[p]\leftarrow 1; break$
\ENDIF

\IF{$C[p_{next}] \neq 0$}
\STATE $C[p]\leftarrow C[p_{next}]; break$
\ENDIF
\STATE $p_{next}\leftarrow D[p_{next}]$
\ENDWHILE
\ENDFOR
\STATE \textbf{return} C

\end{algorithmic}
\label{group}
\end{algorithm}

\section{Experiment}
To validate the performance of the proposed method, we conduct experiments on four public datasets: MLT, ICDAR-2015, RCTW-17 and SCUT-CTW1500.
\subsection{Datasets}

\textbf{MLT}~\cite{MLT} is a dataset provided for ICDAR 2017 competition on multi-lingual scene text detection and script identification. This dataset is composed of complete scene images from 9 languages representing 6 different scripts. Some languages are labeled in word-level such as English, Bangla, French and Arabic. Others are labeled in line-level such as Chinese, Japanese and Korean. This dataset provides 7,200 images for training, 1,800 images for validating and 9,000 images for testing. We use both training set and validation set to train our model.

\textbf{ICDAR2015}~\cite{karatzas2015icdar} is a dataset provided for ICDAR 2015 competition challenge 4. Each text is labeled on word level. Some texts which are unclear or small are labeled as ``DO NOT CARE''. There are 1,000 images for training and 500 images for testing.

\textbf{RCTW-17}~\cite{shi2017icdar2017} is a competition on reading Chinese Text in images which provides a large-scale dataset that consists of various kinds of images, including street views, posters, menus, indoor scenes, and screenshots. There are 8,034 images for training and 4,229 images for testing. Text lines in this dataset are labeled in line-level.

\textbf{SCUT-CTW1500}~\cite{yuliang2017detecting} is a curved text dataset. Different from previous text detection datasets, each text line is labeled with a polygon whose number of vertices is 14. And the evaluation of SCUT-CTW1500 simply follows the PASCAL VOC protocol~\cite{everingham2010pascal} but calculates IoU between the polygons instead of quadrangles. There are 1,000 images for training and 500 images for testing.

\textbf{Evaluation Metrics} For ICDAR2015, MLT and RCTW-17, we use the online evaluation system provided by each dataset. For SCUT-CTW1500, we evaluate the performance by using the evaluation protocol in~\cite{yuliang2017detecting}. 

\subsection{Implementation Details}
We use ResNet-50~\cite{he2016deep} that is pre-trained on ImageNet~\cite{deng2009imagenet} dataset as our backbone. The number of channels of FPN is set to $256$, all upsample operators are the bilinear interpolation rather than time-consuming deconvolution, and the computation in convolutional layer includes convolution, batch normalization~\cite{ioffe2015batch} and ReLU~\cite{nair2010rectified}. The network is trained with stochastic gradient descent (SGD). For simplicity following~\cite{chen2018deeplab,xiao2018unified} we adopt ``poly'' learning rate policy. In particular, learning rate is calculated with $base\_lr \times ( 1 - \frac{iter}{max\_iter} )^{power}$ where $iter$ and $max\_iter$ are number of iterations at present and total number of iterations, $base\_lr$ is initial learning rate which is set to $0.005$ and $power$ is set to $0.9$ in our experiment. And we set weight decay to $0.0001$, momentum to $0.9$ and batch size to 12. The threshold of text center instance score is set to 0.7 and the threshold of text border on TS is set to 0.6. By default, we train our model with TS, TCBP and TCD, but only use TS and TCBP in inference. For MLT and RCTW-17, we only use their respective dataset to train our model. But for ICDAR2015 and SCUT-CTW1500, our model is pre-trained on MLT then fine-tuned on their respective dataset.

To make network more robust, we adopt data augmentation for preventing overfitting, especially for limited datasets. Firstly, we rotate images by a random angle of $\{0^\circ, 90^\circ, 180^\circ, 270^\circ\}$ with a probability of $0.5$. After that a random rotation in range $[-10^\circ, 10^\circ]$ is also applied on images. Next, we randomly crop a $512 \times 512$ image from rotated images whose labels are quadrangle using the rules in SSD~\cite{liu2016ssd}. But for curved polygon labeled dataset we crop images without crossing texts, because cropping may destroy curved shape of polygon. In the training, some text lines which are unclear and fuzzy are labeled as ``DO NOT CARE'', we ignore them by setting loss weight to zero. Besides these, the text lines whose height is smaller than 10 pixels are also ignored. The whole algorithm is implemented in PyTorch 0.4.0~\cite{Pytorch} and we conduct all experiments on a regular workstation whose CPU is Intel(R) Core(TM) i7-7700K and GPU is GeForce GTX 1080Ti.

\subsection{Experiments on MLT}
We combine training set and validation set of MLT to train our model with 180,000 iterations. Firstly, we make controlled experiments to examine how each component affects the model performance. We use the same settings except for each tested component in our experiment and test in single scale by resizing the long side of images to 1800. Then we compare our method with other state-of-the-art methods and also evaluate our model with multiple scales. For multi-scale testing, we resize the long side of images to \{1000, 1800, 2600\} pixels and merge multi-scale results by non-maximum suppression (NMS).

\begin{table}

  \begin{center}

    \begin{tabular}{|c|c|c|c|}
    \hline
        Method& P& R & F  \\
      \hline
      Baseline & 57.42&52.53  &54.86  \\
      Ours(a) &80.92 & 67.69 & 73.72 \\
      Ours(b)  &82.34 &67.33 & 74.08 \\
      Ours(c)  &82.78 & 67.92 & 74.62 \\
      Ours(d)  &\textbf{82.82} & \textbf{68.06}& \textbf{74.72} \\
      \hline

    \end{tabular}

  \end{center}
\caption{Ablation experimental results. ``Baseline'': a semantic segmentation result, ``Ours(a)'': only use TCBP in both training and inference, ``Ours(b)'': use center-border binary map and TCD in training and inference, ``Ours(c)'': use both TCBP and TCD in training and inference, ``Ours(d)'': only use TCBP in inference but model is trained with TCD and TCBP.}
\label{table_abalation}
\end{table}

\begin{figure}[t]
  \begin{center}
       \includegraphics[width=1.\linewidth]{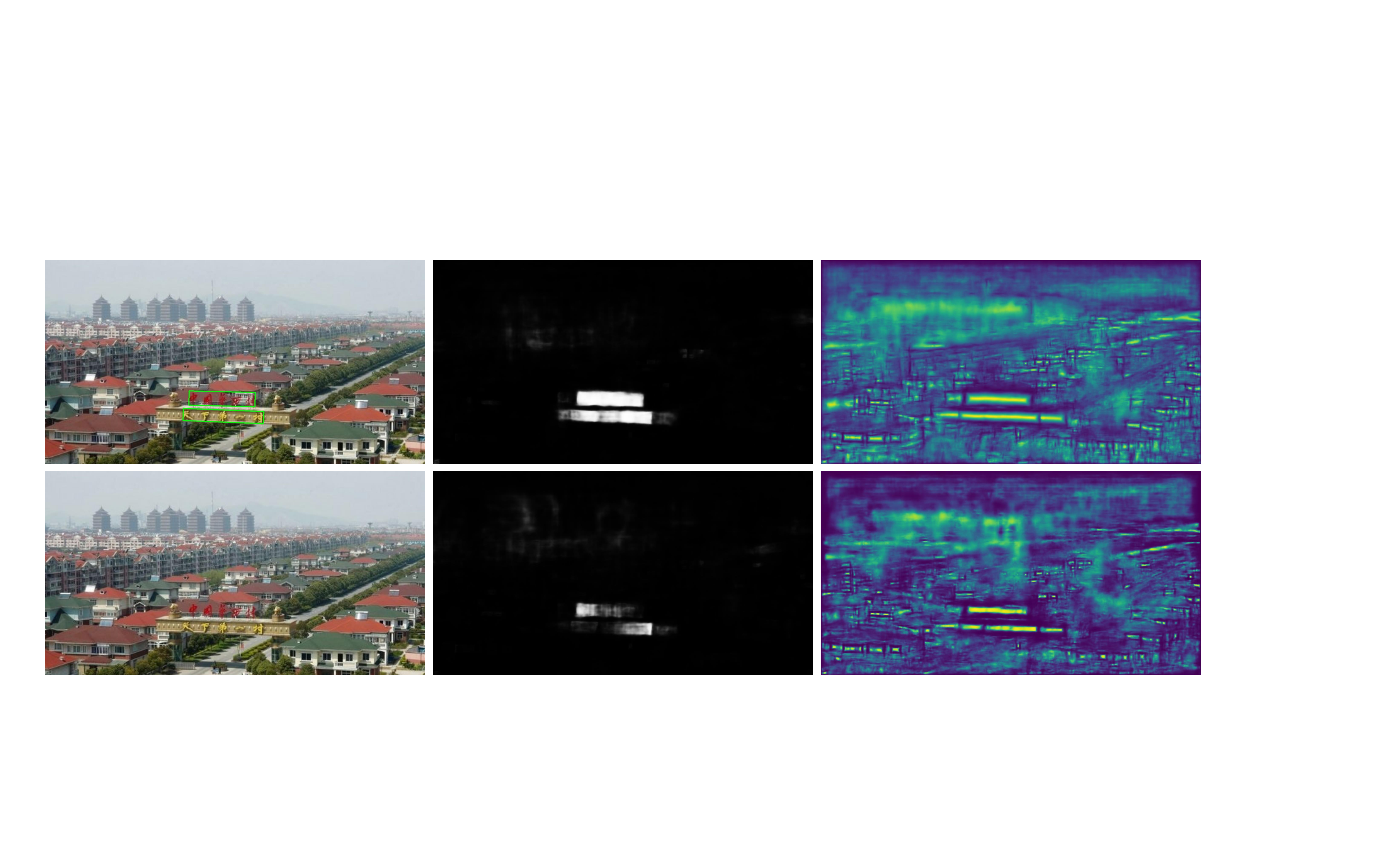}
    \end{center}
    \caption{A comparison of TCBP (Top) and center-border binary classification map (Bottom). From left to right: detection result, TS map and TCBP (or center-border binary classification) map. 
    }
  \label{compare_classes}
\end{figure}

\subsubsection{Ablation Experiments}
\noindent
\textbf{Comparisons with baseline.} Firstly we train a semantic segmentation model named ``Baseline'' which only includes TS. Table~\ref{table_abalation} shows the result of ``Baseline'' where semantic segmentation map does not have enough information to separate adjacent text lines. However, TCBP can well address this problem and its gradient direction can be used in grouping pixels. So the results of ``Ours(a)'' are much better than those of ``Baseline''.

\noindent
\textbf{How important is TCD.}
Although there are enough information in TCBP for both separating and grouping text lines, we find that TCD can make model more robust. In order to prove this, we train a model with both TCD and TCBP. Then using TCBP or TCD to group text lines (``Ours(d)'' vs. ``Ours(c)'' as shown in Table~\ref{table_abalation}). ``Ours(d)'' exceeds ``Ours(a)'' $1\%$ and exceeds ``Ours(c)'' $0.1\%$ which indicates that TCD is not necessary in inference but can help TCBP learning better in the training stage. From Eq.~(\ref{probability_cal}) and Eq.~(\ref{cal_direction}), we can see that TCD and TCBP have a strong correlation, they learn via different expressions of the same feature. TCD can push close pixels belong to the same text line and distant pixels belong to different text lines.

\noindent
\textbf{Probability map vs. binary map.} If there is TCD, we can group text lines without TCBP. The TCBP is only used for separating text which can also be done by center-border binary classification map. To prove TCBP still plays an important role in this case, we train a model named ``Ours(b)'' with center-border binary classification map and TCD. The center-border binary classification map classifies pixels as two classes (center or border), and the center threshold used in label is $\gamma$ which is the same as in TCBP. As shown in Table~\ref{table_abalation}, TCBP (``Ours(c)'') improves center-border binary classification map (``Ours(b)'') with gains of $0.44\%$ on Precision and $0.59\%$ on Recall, which demonstrates that TCBP is more effective compared with center-border binary classification map when combining with TCD. As illustrated in Fig.~\ref{compare_classes}, TCBP is smoother than center-border binary classification map.

\subsubsection{Comparing with other state-of-the-art methods}
We compare our method with other state-of-the-art methods in Table~\ref{table_mlt}. To be fairly comparable, here the result from~\cite{xue2018accurate} is based on ResNet-50. We can observe that our method significantly outperforms other methods. For the single-scale setting, our method achieves the F-measure of $74.72\%$ with an absolute gain of about 8\% over the most competing method in~\cite{lyu2018multi}. For the multi-scale setting, TextMountain achieves the F-measure of $76.85\%$ with an absolute gain of about 4\% over the most competing method in~\cite{lyu2018multi}.

\begin{table}

  \begin{center}

    \begin{tabular}{|l|c|c|c|c|}
      \hline
         Method  & P& R& F\\

      \hline\hline
        linkage-ER-Flow~\cite{MLT} &44.48 & 25.59&32.49 \\
        TH-DL~\cite{MLT} &67.75&34.78&45.97 \\
        SARI\_FDU\_RRPN\_v1~\cite{MLT} &71.17&55.50&62.37 \\
        Sensetime OCR~\cite{MLT} &56.93&69.43& 62.56 \\ 
        SCUT\_DLVClab1\cite{MLT} &80.28&54.54&64.96 \\ 
        Lyu \etal~\cite{lyu2018multi} &\textbf{83.8}&55.6&66.8 \\
        TextMountain &82.82 & 68.06 & 74.72 \\

        Xue \etal MS~\cite{xue2018accurate} &73.9 & 60.6& 66.6\\
        Lyu \etal MS~\cite{lyu2018multi} &74.3&70.6&72.4 \\ 
        TextMountain MS &79.33 & \textbf{74.51} & \textbf{76.85} \\

      \hline

    \end{tabular}

  \end{center}
\caption{Results on MLT. MS means multi-scale.}
\label{table_mlt}
\end{table}

\begin{table}

  \begin{center}

    \begin{tabular}{|l|c|c|c|c|}
      \hline
         Methods  & P& R& F&FPS\\

      \hline\hline
      Zhang \etal~\cite{zhang2016multi} &71&43  &54&-\\
      SegLink~\cite{shi2017detecting} &73.1&76.8   &75.0& -\\
      EAST~\cite{zhou2017east} & 83.57& 73.47  &78.20 & \textbf{13.2} \\
      EAST~\cite{zhou2017east} & 83.27& 78.33  &80.72 & - \\
      He \etal~\cite{he2017deep} & 82 & 80 &81 & -\\


      PixelLink~\cite{deng2018pixellink} & 85.5 & 82.0 &83.7&3.0\\
      Lyu \etal~\cite{lyu2018multi}    &\textbf{89.5}& 79.7& 84.3&1\\
      TextSnake~\cite{long2018textsnake} &84.9&80.4&82.6&1.1\\
      \hline\hline
      TextMountain  & 88.51 &\textbf{84.16} &    \textbf{86.28} & 10.5\\

      \hline

    \end{tabular}

  \end{center}
\caption{Results on ICDAR2015.}
\label{table0}
\end{table}

\begin{table}

  \begin{center}

    \begin{tabular}{|c|c|c|c|c|}
      \hline
       dataset &resolution& C & G  & other\\
       \hline
       IC15 & $1280\times768$ &0.1241& $0.0006$ & 0.0946\\
       RCTW-17 &500 (long)&0.1646&0.0003&0.0268\\
       RCTW-17 &1500 (long)&1.2910&0.0013&0.1653\\

      \hline
    \end{tabular}

  \end{center}
      \caption{The efficiency experiment on ICDAR2015 and RCTW-17. C is the time (in second) consumed in CPU group, G is the time (in second) consumed in GPU group and other is the time (in second) consumed in other stages.}
\label{table_time}
\end{table}

\subsection{Experiments on ICDAR2015}
We validate the performance of our method on ICDAR2015 dataset to evaluate its ability for oriented text. We fine-tune our model 60,000 iterations on ICDAR2015 training set. Following~\cite{deng2018pixellink,lyu2018multi,long2018textsnake} we resize images to $1280 \times 768$ in inference and report the single-scale result. We compare our method with other state-of-the-art methods and show results in Table~\ref{table0}. Our method achieves better performance (precision: 88.51\%, recall: 84.16\% and F-measure: 86.28\%) compared with other segmentation based methods~\cite{deng2018pixellink,lyu2018multi}. Moreover, by acceleration using parallel grouping, TextMountain can run at a speed of 10.5 FPS, is faster than most methods. To explore the efficiency of parallel grouping, we compare two methods, one is implemented in cython based on the code of PixelLink\footnote {\url{https://github.com/ZJULearning/pixel_link}} and the other one is implemented in GPU with Algorithm~\ref{group}. In Table~\ref{table_time}, obviously our grouping using GPU parallel computation is much faster ($200\times$) than cython code.

\begin{table}

  \begin{center}

    \begin{tabular}{|c|c|c|c|c|}
      \hline
         Methods  & P& R& F&FPS\\

      \hline\hline
      Official baseline~\cite{shi2017icdar2017}&  76.03& 40.44 & 52.78 &8.9\\
      EAST-ResNet*~\cite{zhou2017east}&59.7&47.8&53.1& 7.4\\
      RRD~\cite{liao2018rotation} &72.4&45.3&55.7&\textbf{10}\\

      TextMountain &\textbf{80.80}&\textbf{55.24}&\textbf{65.63}& 6 \\
      \hline\hline
      RRD MS~\cite{liao2018rotation}&\textbf{77.5}&59.1&67.0&-\\
      Xue \etal MS~\cite{xue2018accurate} &74.2&58.5  & 65.4&-\\
      TextMountain MS&76.82&\textbf{60.29}&\textbf{67.56}& - \\

      \hline

    \end{tabular}

  \end{center}
\caption{Results on RCTW-17. (* indicate the result is from~\cite{liao2018rotation}).}
\label{table_rctw}
\end{table}

\begin{figure*}
  \begin{center}
       \includegraphics[width=0.8\linewidth]{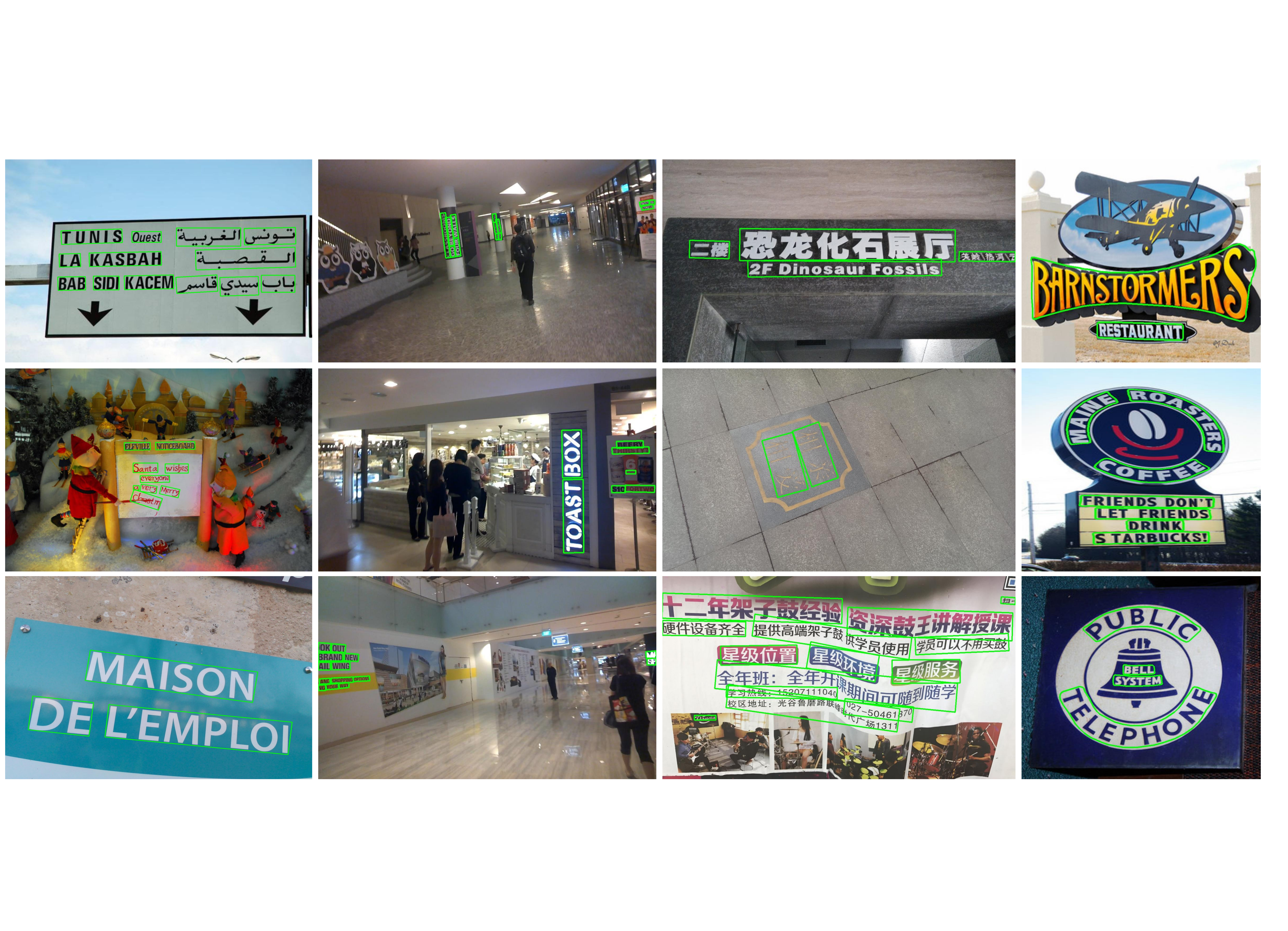}
    \end{center}
    \caption{Qualitative results by the proposed method. From left to right: MLT, ICDAR2015, RCTW-17 and SCUT-CTW1500. 
    }
  \label{results}
\end{figure*}

\subsection{Experiments on RCTW-17}
We validate the performance of our method on RCTW-17 dataset to evaluate its ability for long Chinese text. We train our model with 8,034 images from RCTW-17 for 180,000 iterations. Results are shown in Table~\ref{table_rctw}. For a fair comparison, the result from~\cite{xue2018accurate} is based on ResNet-50. Firstly, we test our model in the single-scale setting, we resize the long side of images to 1500. Compared with regression based methods (EAST, RRD), our method achieves much better performance on high resolution images, because the receptive field of our method does not need to cover the whole text line. In the multi-scale setting, the long side of images is resized to \{500, 1000, 1500, 2000, 2500\}. The F-measure of TextMountain is 67.56\%, yielding comparable results with other methods. In Table~\ref{table_time}, we also evaluate the efficiency of grouping. As the size of the image increases, CPU grouping consumes plenty of time. For example, when the long side of image is 1500, CPU grouping consumes 1.2910s for one image, but our GPU grouping only needs 0.0013s. It proves the proposed algorithm is quite efficient for high resolution images.

\begin{table}

  \begin{center}

    \begin{tabular}{|l|c|c|c|}
      \hline
         Method  & P& R& F\\

      \hline\hline
            Seglink*~\cite{shi2017detecting}&42.3  & 40.0&40.8\\
            SWT*~\cite{epshtein2010detecting} & 20.7&9.0&12.5\\
            CTPN*~\cite{tian2016detecting}&60.4 &53.8 & 56.9\\
            EAST*~\cite{zhou2017east}&78.7 &49.1&60.4\\
            DMPNet*~\cite{liu2017deep}& 69.9&56.0 &62.2\\
            CTD+TLOC~\cite{yuliang2017detecting}  & 77.4& 69.8  & 73.4\\
            TextSnake~\cite{long2018textsnake} &67.9&\textbf{85.3}&75.6\\

      \hline\hline
      TextMountain  & \textbf{82.9} &83.4 & \textbf{83.2}\\

      \hline

    \end{tabular}

  \end{center}
      \caption{Results on SCUT-CTW1500. (* indicate the result is from~\cite{yuliang2017detecting}).}
  \label{table_ctd}
\end{table}

\subsection{Experiments on SCUT-CTW1500}
On SCUT-CTW1500, we evaluate the performance of our method for detecting curved text lines. We fine-tune our model 60,000 iterations on SCUT-CTW1500 training set and resize the long side of images to 800 in inference. For fairness, we report our single-scale result. The results are listed in Table~\ref{table_ctd}. The proposed method achieves much better results (82.9\%, 83.4\%, 83.2\% in precision, recall and F-measure) compared with other methods which proves that TextMountain can handle well with curved text lines.

\section{Conclusion and Future Work}
In this study, we propose a novel text detection method named TextMountain. The proposed method achieves state-of-the-art or comparable performance on both traditional quadrangle labeled datasets (MLT, ICDAR2015, RCTW-17) and newly-released polygon labeled curved dataset (SCUT-CTW1500). From Fig.~\ref{results}, we can observe that our method is robust to the variation of shape, angle and size. And benefitting from the parallel grouping, the proposed method is also efficient. In the future, we will explore the end-to-end system for robust text detection and recognition. 

{\small
\bibliographystyle{ieee}
\bibliography{egbib}
}

\clearpage

\appendix

\section{Appendix}
We illustrate the detail of curved text line label calculation in this supplementary material.
\subsection{Label Calculation for Curved Text Line}

\begin{figure}[t]
  \begin{center}
       \includegraphics[width=0.8\linewidth]{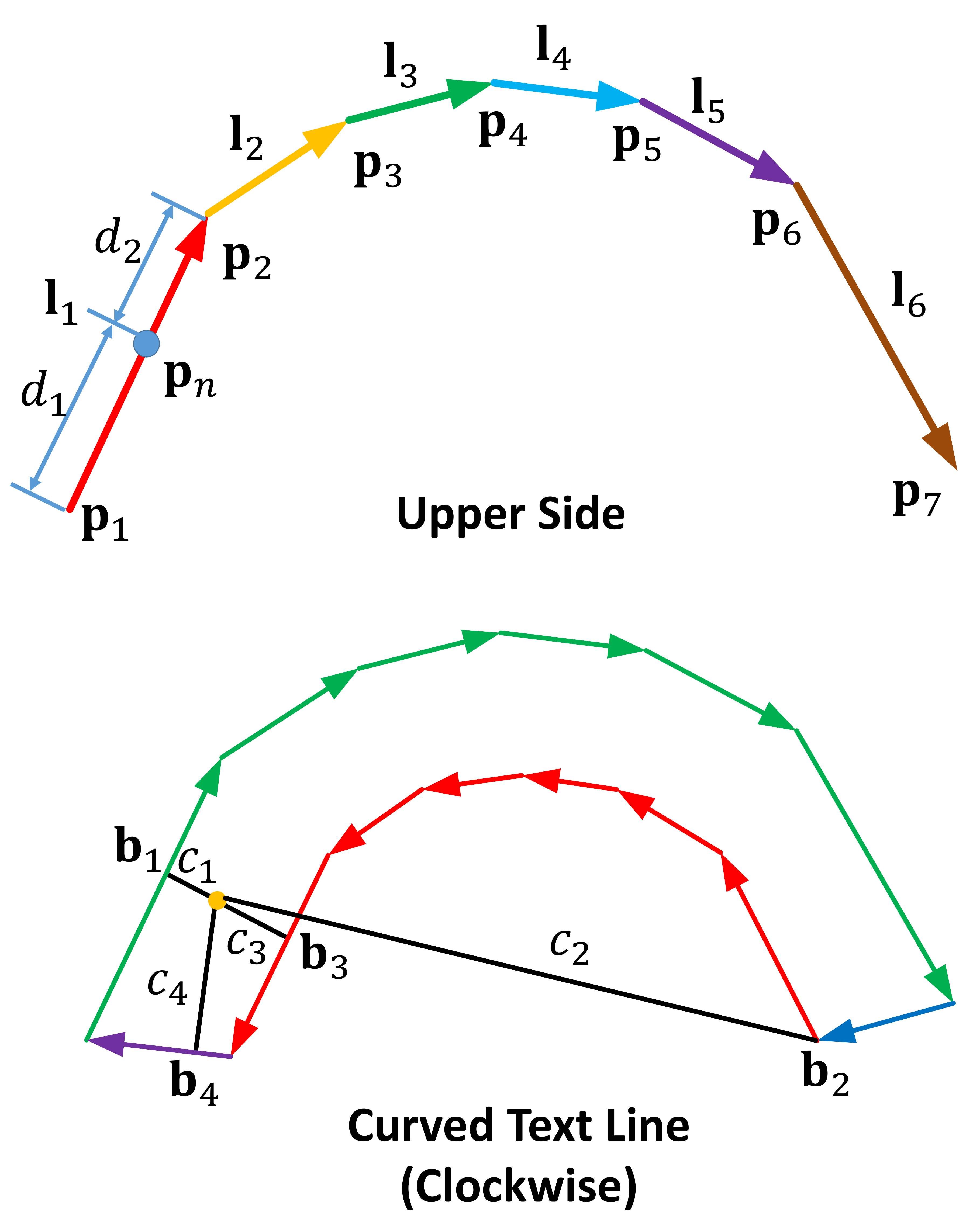}
    \end{center}
    \caption{Bottom: curved text line. Top: the upper side of text line. Each line or side is in a different color for better visualization.
    }
  \label{dis_cal_curve}
\end{figure}

We illustrate the detail of our label calculation on SCUT-CTW1500~\cite{yuliang2017detecting} and the label rule can easily be extended to other curved text datasets. SCUT-CTW1500 labels each text with 14 vertexes and seven of them form a curved line. As shown in Fig. \ref{dis_cal_curve} (bottom), curved text line also has four sides, but two of them may be curved and the direction of lines is clockwise. Actually, the curved line is smooth and angle of line is gradually changing, labeling with finite points will lead to mutation, so we firstly smooth angle of line before calculating label. Taking one curved side as an example, one side is labeled with 7 points and 6 lines in Fig. \ref{dis_cal_curve} (top). $\mathbf{l}_{i}$ is the $i$-th line, $\mathbf{p}_{i}$  and $\mathbf{p}_{i+1}$ are the start point and the end point of $\mathbf{l}_{i}$. $\mathbf{f}_{\mathbf{l}_{i}}$ is the $i$-th line's unit vector and $\mathbf{f}_{\mathbf{p}_{i}}$ is the unit vector on the $i$-th point, we use the mean of point's adjacent lines as its value which can be formulated as:
\begin{equation}
\mathbf{f}_{\mathbf{p}_{i}}=
\begin{cases}
\mathbf{f}_{\mathbf{l}_{1}}&i=1\\
\mathbf{f}_{\mathbf{l}_{6}}&i=7\\
\frac{\mathbf{f}_{\mathbf{l}_{i-1}}+\mathbf{f}_{\mathbf{l}_{i}}}{\|\mathbf{f}_{\mathbf{l}_{i-1}}+\mathbf{f}_{\mathbf{l}_{i}}\|}& \text{otherwise}
\end{cases}
\end{equation}

and other points' unit vectors are calculated via bilinear interpolation. Taking $\mathbf{p}_{n}$ as an example, $\mathbf{p}_{n}$ is between $\mathbf{p}_{i}$ and $\mathbf{p}_{i+1}$, $\mathbf{f}_{\mathbf{p}_{n}}$ is calculated as:
\begin{equation}
\mathbf{f}_{\mathbf{p}_{n}}=\frac{d_{i+1} \mathbf{f}_{\mathbf{p}_{i}}+d_{i} \mathbf{f}_{\mathbf{p}_{i+1}}}{\|d_{i+1} \mathbf{f}_{\mathbf{p}_{i}}+d_{i} \mathbf{f}_{\mathbf{p}_{i+1}}\|}
\label{bilinear_inter}
\end{equation}
where $d_{i}$ is the distance of $\mathbf{p}_{n}$ to $\mathbf{p}_{i}$, and the center-direction angle of $\mathbf{p}_{n}$ is:
\begin{equation}
\theta_{\mathbf{p}_{n}}=\angle(\mathbf{f}_{\mathbf{p}_{n}})+\frac{\pi}{2}
\label{angle}
\end{equation}
where $\angle(\mathbf{f})$ represents the angle of vector $\mathbf{f}$, and we rotate the vector $\frac{\pi}{2}$ clockwise make it point to center.
\begin{figure*}
  \begin{center}
       \includegraphics[width=1.0\linewidth]{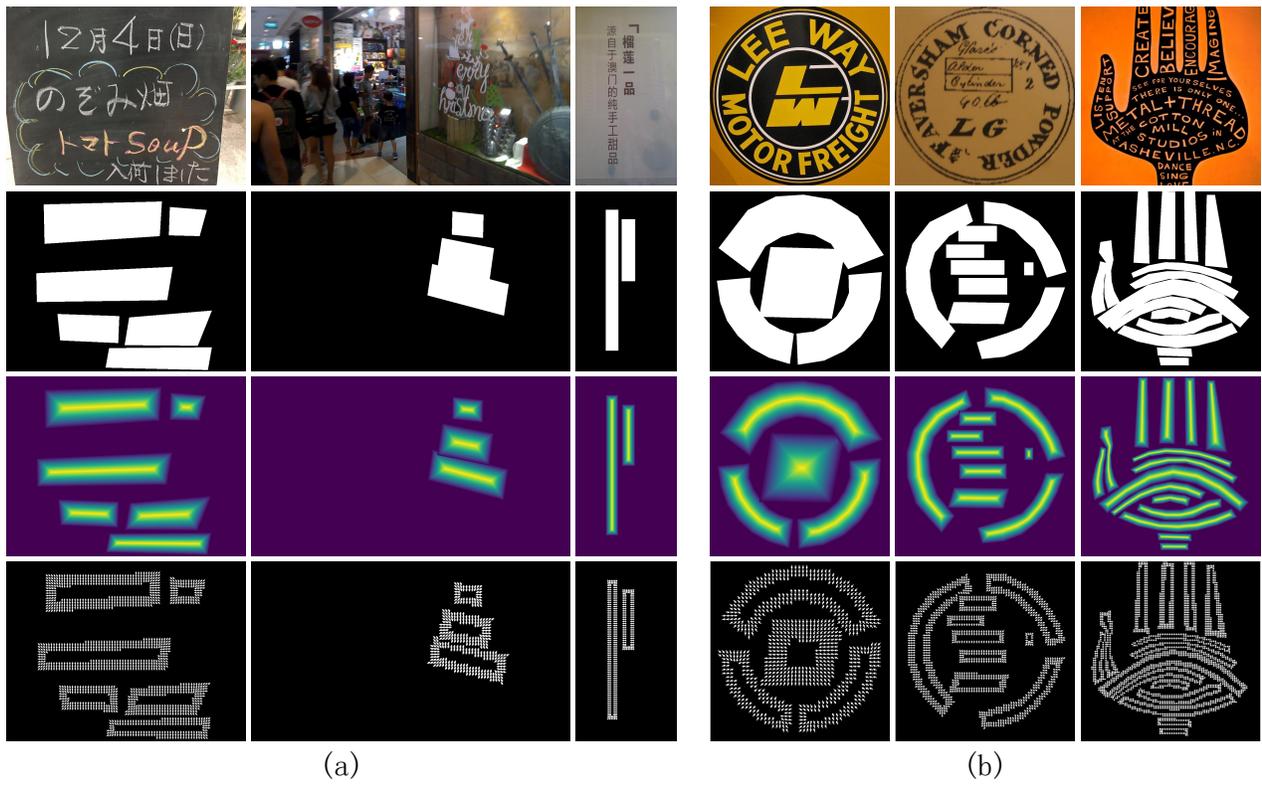}
    \end{center}
    \caption{Quadrangle text line's label (a) and curved text line's label (b). From top to down: original image, TS, TCBP and TCD. We only show TCD on border region for better visualization. 
    }
  \label{label_show}
\end{figure*}

Now we begin calculating labels. As shown in Fig. \ref{dis_cal_curve} (bottom), we calculate the closest point from side to calculated point then mark it as $\mathbf{b}_{i}$, the TCD is same as before, height of text can be formulated as:
\begin{equation}
h_{\mathbf{x}}= \min(c_{1}+c_{3},c_{2}+c_{4})
\label{height_cal_curve}
\end{equation}
where $\mathbf{x}$ is the calculated point, $c_{i}$ is the distance of point $\mathbf{b}_{i}$ to point $\mathbf{x}$, then the TCBP of $\mathbf{x}$ can be calculated as:
\begin{equation}
TCBP_{\mathbf{x}}=\frac{2 \times \min (c_{1},c_{2},c_{3},c_{4})}{h_{\mathbf{x}}}
\label{probability_cal_curve}
\end{equation}
As for TCD, firstly we calculate $\theta$ of point $\mathbf{b}_{i}$ with Eq. (\ref{bilinear_inter}) and Eq. (\ref{angle}) then mark it as $\theta_{{\mathbf{b}_{i}}}$, TCD can be formulated as:
\begin{equation}
v_{\mathbf{x}}^{1}=\sum\limits_{i=1}^{4}  [\frac{h_{\mathbf{x}}}{2}-c_{i}]_{+} \times \cos(\theta_{{\mathbf{b}_{i}}}) 
\end{equation}
\begin{equation}
v_{\mathbf{x}}^{2}=\sum\limits_{i=1}^{4} [\frac{h_{\mathbf{x}}}{2}-c_{i}]_{+} \times \sin(\theta_{{\mathbf{b}_{i}}})
\end{equation}
\begin{equation}
\mathbf{v}_{\mathbf{x}}=(v_{\mathbf{x}}^{1},v_{\mathbf{x}}^{2})
\end{equation}
where $h_{\mathbf{x}}$ is the height of text line, $[z]_{+}$ represents $\max(z,0)$, $\mathbf{v}_{\mathbf{x}}$ is center-direction vector, we norm it as before:
\begin{equation}
\mathbf{u}_{\mathbf{x}}=\frac{\mathbf{v}_{\mathbf{x}}}{\parallel \mathbf{v}_{\mathbf{x}} \parallel}
\label{cal_direction_curve}
\end{equation}

In order to better illustrate our label rule, we show some labeled examples of MLT~\cite{MLT}, ICDAR2015~\cite{karatzas2015icdar}, RCTW-17~\cite{shi2017icdar2017} and SCUT-CTW1500~\cite{yuliang2017detecting} in Fig.~\ref{label_show}. As shown, the center-border probability map gradually decays from center to border, the center-direction points to center and gradually rotates as pixel moves.

\end{document}